\documentclass{article} %
\usepackage[final]{colm2026_conference}

\usepackage{microtype}
\usepackage{hyperref}
\usepackage{url}
\usepackage{booktabs}
\usepackage{graphicx}
\usepackage{xcolor}
\usepackage{wrapfig}
\usepackage{multirow}
\usepackage{tikz}
\usetikzlibrary{patterns}
\usepackage{enumitem}

\usepackage{lineno}

\definecolor{darkblue}{rgb}{0, 0, 0.5}
\hypersetup{colorlinks=true, citecolor=darkblue, linkcolor=darkblue, urlcolor=darkblue}

\title{
    Train Separately, Merge Together: \\ Modular Post-Training with Mixture-of-Experts
}

\author{%
\textbf{Jacob Morrison}\thanks{Equal contribution.}\hspace{0.4em}$^{1,3}$ \quad \textbf{Sanjay Adhikesaven}$^{* 2,3}$ \quad \textbf{Akshita Bhagia}$^{3}$ \\ \textbf{Matei Zaharia}$^{2}$ \quad \textbf{Noah A. Smith}$^{1,3}$ \quad \textbf{Sewon Min}$^{2,3}$
\\[1ex]
$^1$University of Washington\quad$^2$University of California, Berkeley\quad$^3$Allen Institute for AI\quad
\vspace{.5em} \\
  \texttt{jacobm00@cs.washington.edu} 
  \qquad
  \texttt{sanjay.adhikesaven@berkeley.edu} \\
}

\newcommand{\MODEL}{{BAR}}

\newcommand{\tightparagraph}[1]{
\vspace{-.5em}
\paragraph{#1}
}

\begin{document}

\ifcolmsubmission
\linenumbers
\fi

\maketitle

\begin{abstract}

Extending a fully post-trained language model with new domain capabilities is fundamentally limited by monolithic training paradigms: retraining from scratch is expensive and scales poorly, while continued training often degrades existing capabilities. We present \MODEL\ (Branch-Adapt-Route), which trains independent domain experts, each through its own mid-training, supervised finetuning, and reinforcement learning pipeline, and composes them via a Mixture-of-Experts architecture with lightweight router training.
Unlike retraining approaches that mix all domains and require full reprocessing for any update (with cost scaling quadratically), \MODEL\ enables updating individual experts independently with linear cost scaling and no degradation to existing domains. At the 7B scale, with experts for math, code, tool use, and safety, \MODEL\ achieves an overall score of 49.1, outperforming retraining without mid-training (47.8), and nearly matching retraining with mid-training (50.5). We further show that modular training provides a structural advantage: by isolating each domain, it avoids the catastrophic forgetting that occurs when late-stage RL degrades capabilities from earlier training stages, while significantly reducing the cost and complexity of updating or adding a domain.
Together, these results suggest that decoupled, expert-based training is a scalable alternative to monolithic retraining for extending language models.

\end{abstract}

\section{Introduction}

Language models have achieved exceptional capabilities across a wide range of domains and tasks, serving as powerful general-purpose foundation models. In many practical scenarios, model developers need to continuously update an already trained model by incorporating newly available data or knowledge. For example, improved versions of datasets may become available after initial training, or developers may wish to add new capabilities to the original model. Updating LMs in these settings remains challenging: retraining from scratch is often prohibitively expensive and requires full access to the original training pipeline, while continued training risks losing previously attained capabilities due to catastrophic forgetting and cross-domain interference.
Moreover, extending a \textbf{\em fully post-trained} model presents additional challenges: simply continuing post-training is often insufficient for strong performance, as mid-training is typically critical for learning new capabilities and establishing strong priors that post-training can refine~\citep{runwal2026prismdemystifyingretentioninteraction}.

In this paper, we propose a \textbf{modular mixture of experts} approach to post-training: rather than training a single model across all domains, we train independent domain experts---each through its own mid-training, supervised finetuning (SFT), and reinforcement learning (RL) pipeline---and compose them into a unified model via a Mixture-of-Experts (MoE) architecture.
This modular design decouples learning for each domain, enabling experts to be developed, added, or upgraded independently without re-training other components.

While modular MoE-based training has been explored in prior work for {\em pre-training}, e.g., BTX~\citep{sukhbaatar2024branchtrainmixmixingexpertllms} and FlexOlmo~\citep{shi2025flexolmoopenlanguagemodels}, we find that they fail to extend to post-training.
For instance, freezing shared (non-FFN) parameters during expert training (as done in \citet{shi2025flexolmoopenlanguagemodels}) significantly degrades performance in our setting.
This likely reflects that, unlike pre-training that primarily updates knowledge representations, post-training requires adapting broader behaviors, including attention patterns and handling new tokens, e.g., \texttt{<thinking>}. We address this by progressively unfreezing shared layers across training stages and composing experts via a combination of weight merging and lightweight router training.

At the 7B scale, we train domain experts for math, code, tool use, and safety and merge them into a single 5-expert MoE. \MODEL\ outperforms all baselines that do not require re-running mid-training from scratch: it exceeds retraining with post-training only (49.1 vs.\ 47.8), BTX (46.7), continual post-training (45.3), and model merging (36.9 and 6.5). It even approaches full re-training with mid-training (50.5).
Importantly, our experiments showcase that \MODEL\ enables adding new experts or upgrading existing ones at constant cost, without affecting other experts---avoiding the quadratic cost of monolithic retraining while preserving existing capabilities.

Together, these results position decoupled, expert-based training as a scalable alternative to monolithic retraining for extending language models. %

\begin{figure*}[t]
\centering

\includegraphics[width=\textwidth]{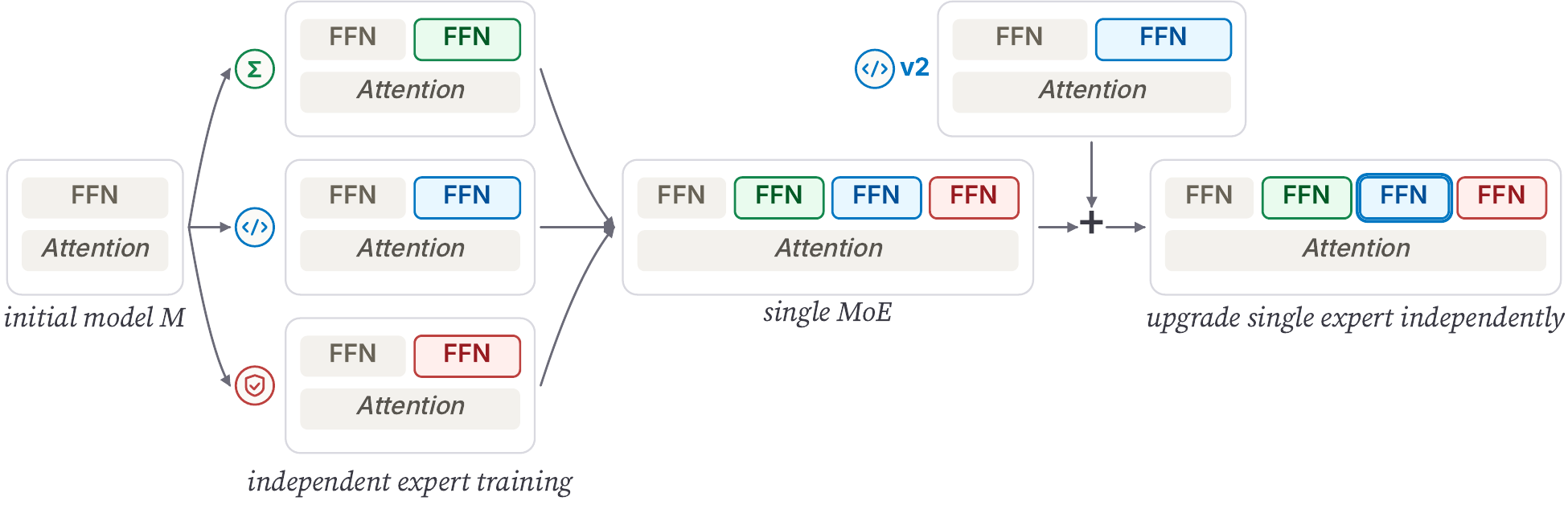}
\vspace{-2.5em}
\caption{Overview of \MODEL. The initial model $M$ (a dense transformer). For each target domain, a two-expert MoE is created: the anchor expert preserves $M$'s capabilities while the domain expert is trained on new data. Each domain follows its applicable pipeline---math and code use the full pipeline (mid-training $\rightarrow$ SFT $\rightarrow$ RLVR), while tool use and safety use SFT only. Shared parameters are progressively unfrozen across stages to minimize divergence between experts. All experts are merged into a single MoE, and a lightweight router is trained on a small sample of SFT data. New experts can later be added (to add a new capability) or swapped in (to upgrade a capability) without retraining previous experts.}
\label{fig:teaser}
\end{figure*}

\section{Related Work}

\paragraph{Model Merging and Modular Training}
A growing body of work explores combining separately trained models by operating directly in parameter space. Task arithmetic~\citep{task_arithmetic} edits model behavior by adding or subtracting task-specific weight vectors, while TIES~\citep{ties} and DARE~\citep{dare} address interference during merging through sign-based and pruning-based strategies. Model soups~\citep{model_soups} average weights of multiple fine-tuned models to improve accuracy. However, these methods assume relatively small parameter shifts from a shared initialization and fail when models diverge substantially, e.g., after continued mid-training on different domains.

A separate line of work takes a more structured modular approach. Branch-Train-Merge~\citep{li2022branchtrainmergeembarrassinglyparalleltraining} trains domain-specific language model branches independently during pre-training and merges them via weight averaging. Branch-Train-MiX (BTX, \citet{sukhbaatar2024branchtrainmixmixingexpertllms}) extends this by composing branches into a Mixture-of-Experts (MoE) model with a trained router, followed by continued joint training. FlexOlmo~\citep{shi2025flexolmoopenlanguagemodels} improves BTX by introducing coordinated expert training, in which the original model is retained and the shared layers are frozen, and only the expert feed-forward parameters are updated. Critically, all of these approaches focus on pre-training; extending them to post-training remains unexplored and, as we show, requires non-trivial modifications.

\tightparagraph{LM Training Pipelines}
Modern LMs are trained in multiple stages: large-scale pre-training on broad corpora, followed by mid-training on domain-specific data to establish strong priors, and post-training including supervised finetuning (SFT), preference optimization~\citep{rafailov2024directpreferenceoptimizationlanguage}, and reinforcement learning with verifiable rewards (RLVR; \citep{lambert2025tulu3pushingfrontiers}). Recent work has shown that mid-training is critical for acquiring new domain capabilities; without it, post-training alone is often insufficient to establish the reasoning primitives needed for strong downstream performance~\citep{runwal2026prismdemystifyingretentioninteraction}. Open-source efforts~\citep{lambert2025tulu3pushingfrontiers, grattafiori2024llama3herdmodels, olmo2025olmo3} have further demonstrated that careful data curation and stage ordering across this full pipeline are essential for strong multi-domain results.
However, sequential multi-domain training within a single pipeline introduces catastrophic forgetting: later stages such as code or math RL can degrade capabilities learned during earlier stages, which may lack RL data entirely.

\tightparagraph{Mixture-of-Experts}
Sparse Mixture-of-Experts (MoE) architectures~\citep{fedus2022switchtransformersscalingtrillion, lepikhin2020gshardscalinggiantmodels} use a learned router to activate a subset of expert modules per input, enabling increased model capacity without proportional compute costs. This approach has been adopted at scale in both open and closed models~\citep{jiang2024mixtralexperts, deepseek_moe}. While MoE is typically motivated by compute efficiency, we instead leverage MoE as a \textit{composition mechanism}: independently trained domain experts are merged into a unified model, with a lightweight router trained to coordinate them.

\section{\MODEL\: Modular post-training with Mixture-of-Experts} \label{sec:method}

\subsection{Problem Setup}

Throughout this paper, we use \textit{domain} to refer to a target capability (e.g., math, code) defined by a curated data distribution designed to develop that capability.
Given an initial model $M$---in this work, a fully post-trained dense base model---and new domain datasets $\{\mathcal{D}_1, \mathcal{D}_2, \dots, \mathcal{D}_k\}$ corresponding to capabilities to add or improve, \MODEL\ produces domain expert modules $\{E_1, E_2, \ldots, E_k\}$, each trained independently, that are composed into a single model $M' = \textsc{Compose}(M, E_1, \ldots, E_k)$ such that:
\begin{enumerate}[leftmargin=17pt, topsep=0pt,itemsep=1pt]
    \item $M'$ substantially outperforms $M$ on each target domain $\mathcal{D}_i$, and ideally, $M'$ approaches or matches the performance of $E_i$ on $\mathcal{D}_i$.
    \item $M'$ preserves the general capabilities of the base model $M$.
\end{enumerate}
Each expert is trained through a pipeline tailored to its target capabilities, drawing from three stages: mid-training on large-scale domain corpora, supervised finetuning (SFT), and reinforcement learning with verifiable rewards (RLVR). Not all capabilities require every stage; some benefit from mid-training to establish priors that post-training alone cannot provide~\citep{runwal2026prismdemystifyingretentioninteraction}, while some can be acquired through SFT/RLVR alone.

This formulation enables modular upgrades (replacing a single expert without retraining others), avoids catastrophic forgetting (each domain's pipeline is isolated), and supports parallel development across teams. Because each expert is trained independently, the cost of adding new domains scales linearly rather than requiring full retraining over all domains.

\subsection{Model Architecture}

In this work, the base model $M$ is a standard dense transformer, in which each layer consists of a self-attention block followed by a feed-forward network (FFN). \MODEL\ converts $M$ into a Mixture-of-Experts (MoE) architecture~\citep{fedus2022switchtransformersscalingtrillion, lepikhin2020gshardscalinggiantmodels} by replacing each layer's single FFN with multiple parallel FFN ``experts'' and a learned router that determines which experts process each input token.

In the final, fully trained \MODEL\ architecture (Figure~\ref{fig:teaser}), each transformer layer contains: (1) \textbf{shared layers}---the embeddings, attention, layer normalization, and language modeling head---which are common across all experts, and (2) \textbf{expert FFNs}---$k+1$ parallel feedforward networks, one per domain plus an \textit{anchor expert} that preserves $M$'s capabilities.

To construct \MODEL\ from the dense base model $M$, the anchor expert's FFN weights are initialized from $M$'s post-trained FFN parameters and frozen, preserving $M$'s original capabilities. Each domain expert's FFN weights are initialized from the pre-trained (not post-trained) base model, providing a clean starting point for domain-specific training without the behavioral constraints of post-training.

\subsection{Training Algorithms}

\MODEL\ training proceeds in three stages:  %

\tightparagraph{Stage 1: Independent expert training}
Each domain expert $E_i$ is trained independently as a two-expert MoE. Each expert follows the full training pipeline as applicable to its domain:
\begin{enumerate}[leftmargin=17pt, topsep=0pt,itemsep=1pt]
    \item Mid-training (sometimes called annealing, using a decaying learning rate schedule), which is continual pre-training on large-scale corpora to establish strong domain priors,
    \item Supervised finetuning (SFT), which adapts the expert to follow instructions and produce well-formatted outputs for the target domain, and
    \item Reinforcement learning with verifiable rewards (RLVR), which further improves performance on tasks where correctness can be automatically verified.
\end{enumerate}
Each domain expert is an independent MoE, enabling fully decoupled training: experts can be trained in parallel, on different hardware, and on different timelines.

\tightparagraph{Key difference from pre-training}
Prior modular {\em pre-training} approaches such as FlexOlmo~\citep{shi2025flexolmoopenlanguagemodels} freeze all shared layers and train only the FFN. For post-training, however, directly applying this strategy significantly degrades domain experts' performance. We hypothesize this gap arises from a key difference in objectives: pre-training primarily updates knowledge representations (well captured by FFNs), while post-training requires adapting broader behaviors, including attention patterns and handling new tokens, e.g., \texttt{<thinking>}. We therefore adapt the recipe for post-training as follows:
\begin{itemize}[leftmargin=17pt, topsep=0pt,itemsep=1pt]
  \item During mid-training, we follow prior work and freeze all shared layers.
  \item During SFT, embeddings and the LM head are unfrozen to support new tokens.
  \item During RLVR, all shared parameters are unfrozen for the behavioral shifts RL requires.
\end{itemize}
Ablations in \S\ref{sec:changes} show that these modifications are critical for effective post-training.

\tightparagraph{Stage 2: Expert merging}
After independent training, the $k$ domain experts are combined with the anchor expert into a single $(k+1)$-expert MoE. For shared parameters that may have diverged across expert training runs (e.g., attention layers that were unfrozen during RL), \MODEL\ averages the diverged parameters across all domain expert models. We find that averaging and lightweight router training introduces little to no measurable performance loss on domain-specific evaluations compared to any single expert (Table~\ref{tab:expert_interference}). %

\tightparagraph{Stage 3: Router training}
After merging, only the router parameters are trained while all expert and shared weights are frozen. The router is a small linear layer at each transformer block that produces a probability distribution over experts for each input token. Training uses a stratified sample of SFT data from all domains, ensuring the router sees examples from every domain. We find that only 5\% of the full SFT dataset is sufficient for effective router training, making this stage computationally inexpensive (\S\ref{sec:results}).

\section{Experimental Setup}

We design our setup to mimic realistic iterative LM development. We start with an existing fully post-trained model and extend it with four new domains (math, code, tool use, and safety) using newer datasets that were not available when the initial model was trained.

\subsection{Initial Model}

Our initial model $M$ is a 7B-parameter dense transformer based on Olmo~2 \citep{olmo2} that has been {\it fully post-trained} for general-purpose chat and instruction-following capabilities. The model was first pretrained on a broad web corpus, then mid-trained on a general-purpose mixture of web text, StackExchange, Wikipedia, and Flan data~\citep{shi2025flexolmoopenlanguagemodels}, followed by SFT on chat and instruction-following data~\citep{lambert2025tulu3pushingfrontiers, olmo2025olmo3} and RLVR for precise instruction following.

Importantly, this initial model's pre-training data already includes some math and code content (e.g., OpenWebMath, Algebra Stack, and StackExchange), but these are older datasets that have since been surpassed by higher-quality alternatives. This mirrors the realistic setting where a deployed model has baseline capabilities in these domains but would benefit from upgrading to newer data and training methods. %

\subsection{New Domain Data}

We add four new domain experts to the initial model: \textbf{math}, \textbf{code}, \textbf{tool use}, and \textbf{safety}. For each domain, we use newer, higher-quality datasets that became available after the initial model was trained, reflecting the realistic scenario of upgrading a deployed model with improved data. Full dataset details and citations are in Appendix~\ref{app:dataset_details}.

\tightparagraph{Math (mid-training $\rightarrow$ SFT $\rightarrow$ RLVR)} For mid-training, we use the math mid-training mix from \citet{shi2025flexolmoopenlanguagemodels}, which combines Dolmino Math~\citep{olmo2} and FineMath~\citep{finemath}. SFT uses math-specific instruction data from Tulu~3~\citep{lambert2025tulu3pushingfrontiers}, mixed with general SFT data to prevent catastrophic forgetting of non-math capabilities (see Table~\ref{tab:sft_mixing}). RLVR uses the math verifiable reward data from OLMo~3~\citep{olmo2025olmo3}.

\tightparagraph{Code (mid-training $\rightarrow$ SFT $\rightarrow$ RLVR)} We experiment with two code expert versions: v1 uses StarCoder-based~\citep{li2023starcodersourceyou} mid-training data and code SFT data from Tulu~3~\citep{lambert2025tulu3pushingfrontiers}, while v2 uses the OLMo~3~\citep{olmo2025olmo3} code mid-training data, SFT data, and RLVR data. As with math, SFT mixes code-specific instruction data with general SFT data (Table~\ref{tab:sft_mixing}). Comparing v1 and v2 allows us to evaluate modular upgrades (Table~\ref{tab:modular_upgrade}).

\tightparagraph{Tool use (SFT only)} Tool use involves learning to generate structured function calls and process their outputs. Unlike math and code, we train a tool use expert using only SFT, with the Dolci tool use data from OLMo~3~\citep{olmo2025olmo3}, mixed with general SFT data.

\tightparagraph{Safety (SFT only)} Safety training teaches the model to refuse harmful requests while avoiding over-refusal of benign ones. We train using only SFT, with safety data drawn from CoCoNot~\citep{coconot}, WildGuardMix~\citep{han2024wildguardopenonestopmoderation}, and WildJailbreak~\citep{jiang2024wildteamingscaleinthewildjailbreaks}, following OLMo~3~\citep{olmo2025olmo3}, mixed with general SFT data.

Training details and hyperparameters for all stages are in Appendix~\ref{app:training_details}.

\subsection{Evaluations}

We evaluate across 19 benchmarks spanning 7 categories: chat (AlpacaEval~\citep{alpacaeval}, IFEval~\citep{zhou2023instructionfollowingevaluationlargelanguage}), knowledge (MMLU~\citep{hendrycks2021measuringmassivemultitasklanguage}, PopQA~\citep{mallen2023trustlanguagemodelsinvestigating}, SimpleQA~\citep{wei2024measuringshortformfactualitylarge}), reasoning (BBH~\citep{suzgun2022challengingbigbenchtaskschainofthought}, GPQA~\citep{rein2023gpqagraduatelevelgoogleproofqa}, ZebraLogic~\citep{lin2025zebralogicscalinglimitsllms}, AGIEval~\citep{zhong2023agievalhumancentricbenchmarkevaluating}), math (MATH~\citep{hendrycks2021measuringmathematicalproblemsolving}, GSM8K~\citep{gsm8k}), code (HumanEval+, MBPP+;~\citealp{liu2023codegeneratedchatgptreally}), tool use (BFCL~\citep{patil2025bfcl}), and safety (HarmBench~\citep{mazeika2024harmbenchstandardizedevaluationframework}, TrustLLM~\citep{huang2024trustllmtrustworthinesslargelanguage}, WildGuard~\citep{han2024wildguardopenonestopmoderation}, WildJailbreak~\citep{jiang2024wildteamingscaleinthewildjailbreaks}, DAN~\citep{shen2024donowcharacterizingevaluating}). Evaluation details are provided in Appendix~\ref{app:detailed_results}.
We report category-level averages in the main paper, with per-benchmark results in Appendix~\ref{app:detailed_results}. The overall score is the unweighted average across all seven category averages.

\subsection{Baselines}

We compare against six baselines: (1)~\textbf{Continual post-training}, which continues training the initial model on all new domain data sequentially without mid-training; (2,3)~\textbf{Dense model merging} with and without mid-training, which trains independent dense models and merges them via weight averaging~\citep{task_arithmetic, model_soups, morrison2024mergelearnefficientlyadding}; (4)~\textbf{BTX}~\citep{sukhbaatar2024branchtrainmixmixingexpertllms}, which trains five fully independent dense models on domain-specific data and combines them into a 5-expert MoE with router training; (5)~\textbf{Re-training (post-train only)}, a dense model re-trained on all domain data through SFT and RLVR without mid-training; and (6)~\textbf{Re-training (mid-train and post-train)}, which re-runs mid-training on all data followed by the full post-training pipeline, representing the upper bound requiring complete access to the original training pipeline.

In terms of development cost, retraining approaches require retraining all domains jointly whenever any single domain is updated, and additionally require tuning cross-domain data mixtures, an iterative process that is quite costly. \MODEL\ trains each expert as a 2-expert MoE, making individual expert training moderately more expensive than BTX or merging baselines which use dense 7B models, but upgrading a single domain requires retraining only that expert and the router, while all other baselines require retraining the entire model.

\section{Results} \label{sec:results}

\subsection{Main Results}

\begin{table}[t]
\begin{center}
\resizebox{\textwidth}{!}{
\setlength{\tabcolsep}{3pt}
\begin{tabular}{lrrrrrrrr}
\toprule
\textbf{Model} & \textbf{Overall} & \textbf{Knowledge} & \textbf{Reasoning} & \textbf{Chat} & \textbf{Math} & \textbf{Code} & \textbf{Tool Use} & \textbf{Safety} \\
\midrule
\textit{Initial model $M$ (7B)} & 31.3 & 28.5 & 29.8 & \textbf{48.9} & 23.6 & 11.8 & 25.3 & 51.3 \\
\midrule
\multicolumn{9}{l}{\textit{Math expert (2$\times$7B)}} \\
\quad + SFT                      & 36.8 & 28.8 & \underline{31.2} & 40.9 & 41.9 & 20.5 & 21.6 & 72.7 \\
\quad + RL                       & 39.3 & \underline{29.0} & 30.8 & 42.5 & 55.8 & 22.1 & 19.8 & 75.4 \\
\midrule
\multicolumn{9}{l}{\textit{Code expert (2$\times$7B)}} \\
\quad + SFT                      & 38.5 & 28.8 & 29.1 & 40.1 & 25.5 & 49.3 & 19.7 & 77.3 \\
\quad + RL                       & 38.8 & 28.5 & 29.2 & 41.0 & 26.9 & \textbf{50.4} & 19.8 & 75.3 \\
\midrule
\multicolumn{9}{l}{\textit{Tool use expert (2$\times$7B)}} \\
\quad + SFT                         & 37.2 & 28.5 & 28.7 & 39.3 & 21.8 & 16.9 & \underline{46.4} & 79.1 \\
\midrule
\multicolumn{9}{l}{\textit{Safety expert (2$\times$7B)}} \\
\quad + SFT                         & 35.6 & 28.7 & 28.8 & 38.1 & 22.4 & 15.7 & 21.1 & \textbf{94.6} \\
\midrule
\multicolumn{9}{l}{\textit{Combined (\textbf{Ours})}} \\
\MODEL\ 5$\times$7B (all experts)  & \textbf{49.1} & 28.4 & 30.8 & 38.7 & \underline{56.2} & \underline{49.9} & 45.6 & \underline{94.0} \\
\midrule
\multicolumn{9}{l}{\textit{Baselines}} \\
Continual post-training          & 45.3 & 26.8 & 29.4 & 38.8 & 45.0 & 42.9 & 40.9 & 93.1 \\
Model merging (w/ mid-train)     & 6.5 & 0.1 & 10.2 & 5.9 & 0.3 & 0.4 & 19.7 & 9.1 \\
Model merging (w/o mid-train)    & 36.9 & \textbf{29.2} & 30.0 & 42.6 & 32.9 & 25.1 & 19.7 & 78.5 \\
BTX 5$\times$7B                  & 46.7 & 23.9 & 30.6 & 36.4 & \textbf{62.1} & 32.1 & \textbf{47.9} & 93.9 \\
Re-training (post-train only)    & \underline{47.8} & 28.6 & \textbf{31.3} & \underline{43.9} & 48.7 & 43.6 & 45.3 & 92.8 \\
\midrule
Re-training (mid+post-train)$^\dagger$     & 50.5 & 27.0 & 31.8 & 43.2 & 55.9 & 59.6 & 45.9 & 90.4 \\
\bottomrule
\end{tabular}
}
\end{center}
\vspace{-.8em}
\caption{Main results and per-expert performance across training stages. Each domain expert is trained as a 2$\times$7B MoE (anchor expert + domain expert). \textbf{Bold} and \underline{underline} indicate best and second-best across all models (excluding $\dagger$). $\dagger$Re-training with mid-training requires complete access to the original pre-training checkpoint and reprocessing all mid-training data from scratch. Per-benchmark results are in Table~\ref{tab:detailed_results}.}
\label{tab:stage_ablation}
\end{table}

We compare our modular approach against several baselines in Table~\ref{tab:stage_ablation}. Our full \MODEL\ 5$\times$7B achieves an overall score of 49.1, exceeding the retraining (post-train only) baseline (47.8) while achieving substantially higher math performance (56.2 vs.\ 48.7) and higher code performance (49.9 vs.\ 43.6), while maintaining strong safety (94.0 vs.\ 92.8)---consistent with the hypothesis that modular training avoids the catastrophic forgetting that occurs when late-stage RL on math and code degrades safety capabilities learned during earlier SFT.

\MODEL\ also outperforms continual post-training (45.3), which simply continues training the initial dense model on all new domain data without mid-training, confirming that post-training alone is insufficient for strong domain capabilities. The BTX baseline, which trains five dense domain-specific models through the same per-domain recipes and combines them via router training, achieves 46.7 overall, lower than our approach despite using the same domain data and training stages. We attribute this gap to BTX training each expert as a fully independent dense model without shared parameters, which leads to greater divergence and makes composition via routing more difficult. BTX is stronger on math (62.1 vs.\ 56.2) and tool use (47.9 vs.\ 45.6), whereas \MODEL\ is stronger on knowledge (28.4 vs.\ 23.9), chat (38.7 vs.\ 36.4), and code (49.9 vs.\ 32.1). We attribute this to \MODEL\'s shared expert, which restricts divergence between task-specific models, while BTX's fully independent experts can achieve higher individual peak performance and combine less cleanly. \MODEL\ also degrades more gracefully when fewer experts are active (Table~\ref{tab:active_experts_comparison}).

Dense model merging after mid-training fails catastrophically (6.5 overall), producing a nearly non-functional model. We hypothesize that this is because mid-training causes the models to diverge substantially, making merging more difficult. Without mid-training, merging performs better (36.9) but falls short of other approaches.

Re-training with mid-training achieves the highest score (50.5), but requires complete access to the original pre-training checkpoint and reprocessing all data from scratch---impractical for most models and expensive when possible. \MODEL\ achieves competitive performance while enabling modular upgrades and parallel development that retraining cannot support.

\begin{table}[t]
\begin{center}
\resizebox{\textwidth}{!}{
\setlength{\tabcolsep}{4pt}
\begin{tabular}{llrrrrrrrr}
\toprule
\textbf{Model} & \textbf{Experts} & \textbf{Overall} & \textbf{Knowledge} & \textbf{Reasoning} & \textbf{Chat} & \textbf{Math} & \textbf{Code} & \textbf{Tool Use} & \textbf{Safety} \\
\midrule
Initial model $M$ & --- & 31.3 & 28.5 & 29.8 & 48.9 & 23.6 & 11.8 & 25.3 & 51.3 \\
\midrule
\MODEL\ 2$\times$7B & + Math                    & 38.9 & 28.5 & 32.0 & 43.4 & 57.9 & 20.2 & 21.0 & 69.5 \\
\MODEL\ 3$\times$7B & + Code              & 42.3 & 28.4 & 30.4 & 38.5 & 58.5 & 48.1 & 19.9 & 72.4 \\
\MODEL\ 4$\times$7B & + Tool Use                       & 46.5 & 28.7 & 30.7 & 40.6 & 56.5 & 48.3 & 45.8 & 74.9 \\
\MODEL\ 5$\times$7B & + Safety                         & 49.1 & 28.4 & 30.8 & 38.7 & 56.2 & 49.9 & 45.6 & 94.0 \\
\bottomrule
\end{tabular}
}
\end{center}
\vspace{-1em}
\caption{Performance as domain experts are incrementally added to \MODEL. Adding new experts for tool use and safety improves overall performance without degrading math or code, demonstrating that experts do not interfere with each other. Per-benchmark results are in Table~\ref{tab:detailed_interference}.
}
\label{tab:expert_interference}
\end{table}

\begin{table}[t]
\begin{center}
\resizebox{\textwidth}{!}{
\setlength{\tabcolsep}{4pt}
\begin{tabular}{lrrrrrrrr}
\toprule
\textbf{Model} & \textbf{Overall} & \textbf{Knowledge} & \textbf{Reasoning} & \textbf{Chat} & \textbf{Math} & \textbf{Code} & \textbf{Tool Use} & \textbf{Safety} \\
\midrule
\multicolumn{9}{l}{\textit{Math expert (standalone, 2$\times$7B)}} \\
Math expert v1 (SFT only)        & 36.8 & 28.8 & 31.2 & 40.9 & 41.9 & 20.5 & 21.6 & 72.7 \\
Math expert v2 (SFT + RL)        & 39.3 & 29.0 & 30.8 & 42.5 & 55.8 & 22.1 & 19.8 & 75.4 \\
\quad $\Delta$                   & \textcolor{gray}{+2.5} & \textcolor{gray}{+0.2} & \textcolor{gray}{$-$0.4} & \textcolor{gray}{+1.6} & \textcolor{gray}{+13.9} & \textcolor{gray}{+1.6} & \textcolor{gray}{$-$1.8} & \textcolor{gray}{+2.7} \\
\midrule
\multicolumn{9}{l}{\textit{Combined with math upgrade (all other experts held fixed)}} \\
\MODEL\ w/ Math v1               & 47.0 & 28.9 & 31.2 & 38.0 & 43.2 & 49.6 & 45.0 & 92.9 \\
\MODEL\ w/ Math v2               & 49.1 & 28.4 & 30.8 & 38.7 & 56.2 & 49.9 & 45.6 & 94.0 \\
\quad $\Delta$                   & \textcolor{gray}{+2.1} & \textcolor{gray}{$-$0.5} & \textcolor{gray}{$-$0.4} & \textcolor{gray}{+0.7} & \textcolor{gray}{+13.0} & \textcolor{gray}{+0.3} & \textcolor{gray}{+0.6} & \textcolor{gray}{+1.1} \\
\midrule
\midrule
\multicolumn{9}{l}{\textit{Code expert (standalone, 2$\times$7B)}} \\
Code expert v1                   & 35.5 & 28.8 & 29.2 & 38.8 & 23.9 & 29.7 & 21.5 & 76.6 \\
Code expert v2                   & 38.8 & 28.5 & 29.2 & 41.0 & 26.9 & 50.4 & 19.8 & 75.3 \\
\quad $\Delta$                   & \textcolor{gray}{+3.3} & \textcolor{gray}{$-$0.3} & \textcolor{gray}{0.0} & \textcolor{gray}{+2.2} & \textcolor{gray}{+3.0} & \textcolor{gray}{+20.7} & \textcolor{gray}{$-$1.7} & \textcolor{gray}{$-$1.3} \\
\midrule
\multicolumn{9}{l}{\textit{Combined with code upgrade (all other experts held fixed)}} \\
\MODEL\ w/ Code v1              & 46.7 & 28.8 & 31.3 & 36.8 & 56.6 & 33.4 & 45.9 & 94.3 \\
\MODEL\ w/ Code v2              & 49.1 & 28.4 & 30.8 & 38.7 & 56.2 & 49.9 & 45.6 & 94.0 \\
\quad $\Delta$                   & \textcolor{gray}{+2.4} & \textcolor{gray}{$-$0.4} & \textcolor{gray}{$-$0.5} & \textcolor{gray}{+1.9} & \textcolor{gray}{$-$0.4} & \textcolor{gray}{+16.5} & \textcolor{gray}{$-$0.3} & \textcolor{gray}{$-$0.3} \\
\bottomrule
\end{tabular}
}
\end{center}
\vspace{-.6em}
\caption{Modular expert upgrades. Upgrading individual experts---by adding RL (math) or using newer data and RL (code)---improves the combined \MODEL\ without retraining any other experts. $\Delta$ rows show the difference between v1 and v2. All non-upgraded experts are held fixed. Per-benchmark results are in Table~\ref{tab:detailed_upgrades}.}
\label{tab:modular_upgrade}
\end{table}

\tightparagraph{Training stages are additive}
Table~\ref{tab:stage_ablation} additionally shows per-expert performance across training stages. Additional training stages per domain (mid-training $\rightarrow$ SFT $\rightarrow$ RL for math and code) progressively improve domain-specific performance. For example, the math expert improves from 41.9 (SFT only) to 55.8 (SFT + RL). These per-expert gains transfer to the final merged model, demonstrating the value of full domain-specific post-training pipelines within the modular framework.

\begin{wrapfigure}{r}{0.4\textwidth}
\vspace{-2.5em}
\centering
\includegraphics[width=\linewidth]{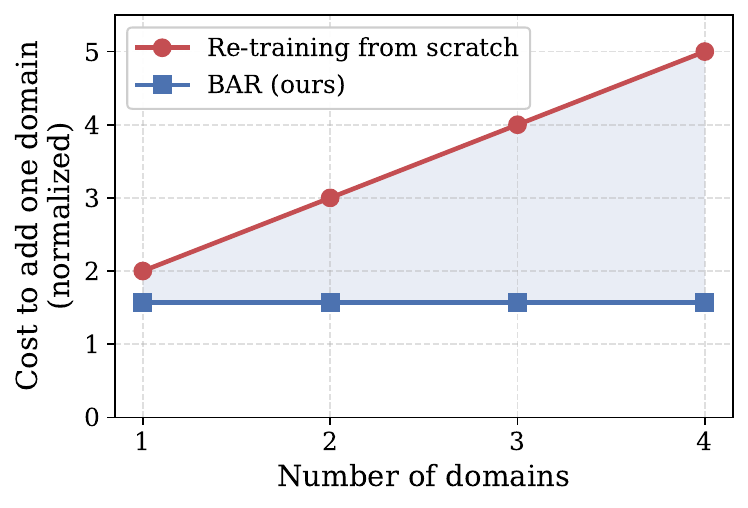}
\vspace{-2em}
\caption{Cost to add each new domain. Re-training must reprocess all domains, so cost grows linearly with the number of domains. \MODEL\ trains only the new expert, keeping cost constant.}
\label{fig:cost_comparison}
\vspace{-9em}
\end{wrapfigure}

\subsection{Adding and Upgrading Experts}

{\rightskip=0.44\textwidth
\tightparagraph{Adding experts} \MODEL\ enables both adding new domain experts and upgrading existing ones without retraining the full model. Table~\ref{tab:expert_interference} shows that new experts can be added incrementally without degrading existing domains: as experts are added from math through safety, overall performance improves while math and code scores remain stable.\par}

\tightparagraph{Upgrading experts} There are two natural ways to upgrade an existing expert: (1) training on newer or higher-quality data, and (2) adding additional training stages such as RL. Table~\ref{tab:modular_upgrade} demonstrates both. For code, replacing an expert trained on older data (v1) with one trained on newer data and RL (v2) improves code performance by +16.5 points while all other domains remain essentially unchanged. For math, adding RL on top of SFT (v1 $\rightarrow$ v2) improves math performance by +13.0 points in the combined model while other domains remain stable.

{\rightskip=0\textwidth
In a non-modular pipeline, upgrading a single domain requires retraining the \textit{entire} model across all domains. With \MODEL\, only the affected expert and router need retraining, and cost scales linearly with the number of domains rather than quadratically (Figure~\ref{fig:cost_comparison}).}

\subsection{Ablations} \label{sec:changes}

\paragraph{Unfreezing shared layers}
\begin{wrapfigure}{r}{0.45\textwidth}
\vspace{-1.5em}
\centering
\resizebox{0.43\textwidth}{!}{
\setlength{\tabcolsep}{3pt}
\begin{tabular}{lrr}
\toprule
\textbf{Model} & \textbf{General} & \textbf{Tool Use} \\
\midrule
Base (no tool use training)      & 42.3 & 19.6 \\
\midrule
Tool use expert (frozen)         & 38.9 & 20.3 \\
Tool use expert (unfrozen)       & 39.3 & 46.4 \\
\bottomrule
\end{tabular}}

\vspace{0.3em}
\includegraphics[width=\linewidth]{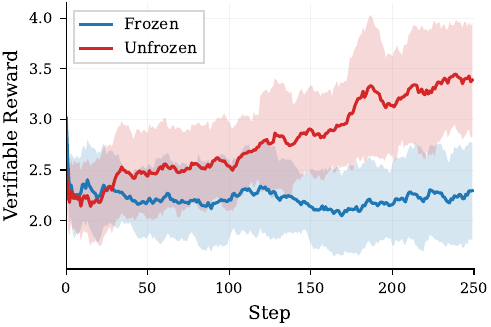}
\vspace{-2em}
\caption{Unfreezing embedding and LM head layers is critical. \textbf{Top:}~Frozen tool use expert fails to learn new tokens (20.3 vs.\ 46.4). \textbf{Bottom:}~For math RL, freezing produces a flat reward curve.}
\vspace{-0.5em}
\label{fig:unfreezing}
\vspace{-1.5em}
\end{wrapfigure}

Prior modular approaches such as FlexOlmo~\citep{shi2025flexolmoopenlanguagemodels} freeze all shared parameters during expert training, which works for pre-training, but we find it does not for post-training. We find that RL requires substantially more unfreezing of shared layers---particularly attention---because RL induces distributional shifts that extend beyond what expert FFNs alone can accommodate (Figure~\ref{fig:unfreezing}). After training, we average all diverged shared layers across experts.

\paragraph{Embedding and language modeling head}
For domains that introduce new special tokens during post-training, such as tool use function-calling formats, the embedding layer and language modeling head must be unfrozen. Figure~\ref{fig:unfreezing} shows that without unfreezing, the tool use expert fails to learn new token representations, producing near-baseline tool use performance (20.3 vs.\ 46.4 with unfreezing).

\paragraph{Mixing general data during SFT preserves general performance}
Training experts on domain-specific SFT data alone causes severe degradation of general capabilities: overall performance drops substantially despite strong in-domain gains (Table~\ref{tab:sft_mixing}). By mixing with the general SFT data used to train the base model, experts preserve their knowledge, reasoning, and chat capabilities while still achieving large domain improvements.

\begin{table}[t]
\begin{center}
\resizebox{\textwidth}{!}{
\setlength{\tabcolsep}{4pt}
\begin{tabular}{llrrrrrrrr}
\toprule
\textbf{Expert} & \textbf{SFT Data} & \textbf{Overall} & \textbf{Knowledge} & \textbf{Reasoning} & \textbf{Chat} & \textbf{Math} & \textbf{Code} & \textbf{Tool Use} & \textbf{Safety} \\
\midrule
Initial model $M$ & --- & 31.3 & 28.5 & 29.8 & 48.9 & 23.6 & 11.8 & 25.3 & 51.3 \\
\midrule
\multirow{2}{*}{Math}      & Domain only          & 25.8   & 22.1 & 23.8   & 15.7 & 33.6 & 22.8 & 19.7 & 43.2 \\
                           & Domain + anchor mix & 36.8 & 28.8 & 31.2 & 40.9 & 41.9 & 20.5 & 21.6 & 72.7 \\
\midrule
\multirow{2}{*}{Code}      & Domain only          & 32.8 & 26.0 & 25.2 & 26.4 & 16.2 & 50.0 & 20.5 & 65.4 \\
                           & Domain + anchor mix & 38.5 & 28.8 & 29.1 & 40.1 & 25.5 & 49.3 & 19.7 & 77.3 \\
\midrule
\multirow{2}{*}{Tool use}  & Domain only          & 28.2 & 18.6 & 20.0 & 21.9 & 12.3 &  6.7 & 44.9 & 73.2 \\
                           & Domain + anchor mix & 37.2 & 28.5 & 28.7 & 39.3 & 21.8 & 16.9 & 46.4 & 79.1 \\
\midrule
\multirow{2}{*}{Safety}    & Domain only          & 31.9 & 25.2 & 24.9 & 19.8 & 20.9 & 16.2 & 19.7 & 96.9 \\
                           & Domain + anchor mix & 35.6 & 28.7 & 28.8 & 38.1 & 22.4 & 15.7 & 21.1 & 94.6 \\
\bottomrule
\end{tabular}
}
\end{center}
\vspace{-.5em}
\caption{Effect of mixing anchor SFT data with domain-specific SFT data. The anchor expert row shows the baseline before domain training. Adding anchor data preserves broad capabilities (knowledge, reasoning, chat) with minimal impact on domain performance. Per-benchmark results are in Table~\ref{tab:detailed_sft_mixing}.}
\label{tab:sft_mixing}
\end{table}

\begin{table}[t]
\begin{center}
\resizebox{\textwidth}{!}{
\begin{tabular}{crrrrrrrr}
\toprule
\textbf{Active Experts} & \textbf{Overall} & \textbf{Knowledge} & \textbf{Reasoning} & \textbf{Chat} & \textbf{Math} & \textbf{Code} & \textbf{Tool Use} & \textbf{Safety} \\
\midrule
1 & 31.1 & 22.8 & 23.4 & 31.0 & 12.2 & 10.3 & 31.6 & 86.5 \\
2 & 38.8 & 27.5 & 28.1 & 37.1 & 23.4 & 19.3 & 44.5 & 91.9 \\
3 & 36.8 & 25.7 & 27.4 & 33.1 & 19.5 & 16.5 & 42.2 & 93.3 \\
4 & 48.2 & \textbf{28.5} & \textbf{31.0} & 37.5 & 55.1 & 48.5 & 42.6 & \textbf{94.5} \\
5 & \textbf{49.1} & 28.4 & 30.8 & \textbf{38.7} & \textbf{56.2} & \textbf{49.9} & \textbf{45.6} & 94.0 \\
\bottomrule
\end{tabular}
}
\end{center}
\vspace{-.7em}
\caption{\MODEL\ 5x7B performance as the number of active experts varies. All models use 5\% router training data with learning rate 1e-4. Best results per domain are \textbf{bolded}.}
\label{tab:active_experts}
\end{table}

\paragraph{Number of active experts}
Performance when activating fewer than all 5 experts during inference is shown in Table~\ref{tab:active_experts}. Activating 4 of 5 experts achieves nearly equivalent performance to the full model, while fewer than 4 leads to substantial drops in math and code.

\section{Discussion}

\paragraph{Structural advantages of modular training}
In standard post-training, all domains pass through every stage sequentially, but not all domains benefit from every stage. For example, safety lacks verifiable RL data, so when math and code RL run later, safety capabilities degrade via catastrophic forgetting. Modular training avoids this by isolating each domain's pipeline. This advantage only became apparent by extending modular training to post-training, where behavioral shifts during SFT and RL require adapting shared layers (\S\ref{sec:changes}), unlike prior work that froze all shared parameters. Beyond quality, modularity provides a cost advantage: adding or upgrading domains scales linearly (one expert per domain), while re-training requires reprocessing all domains each time, scaling quadratically (Figure~\ref{fig:cost_comparison}).

\paragraph{Limitations and future work}
Total parameters grow linearly with the number of experts, increasing inference cost as domains are added. While activating 4 of 5 experts achieves near-full performance (Table~\ref{tab:active_experts}), better routing strategies are needed to reduce cost without sacrificing quality. Upgrading the anchor model currently requires retraining all domain experts. Future directions include scaling to finer-grained experts, extending to new modalities, and upgrading the base model without full expert retraining.

\section{Conclusion}

We have presented \MODEL, a modular approach to post-training that extends fully trained language models with new domain capabilities via independently trained mixture-of-experts. By isolating each domain's training pipeline, \MODEL\ avoids the catastrophic forgetting inherent in multi-domain training, outperforming both a strong dense baseline and Branch-Train-MiX at the 7B scale. Our results demonstrate that domain experts can be upgraded independently, composed without interference, and integrated via lightweight router training, offering a practical path toward scalable, modular LM development.

\section*{Acknowledgements}

We would like to thank Kevin Farhat, Weijia Shi, Parth Asawa, the SM group members, and Ai2 team members for valuable discussions and feedback.

This material is based upon work supported by the National Science Foundation under Award No. 2413244, the National Science Foundation CISE Graduate Fellowships under Grant No. 2313998, and the Office of Naval Research under Award No. N00014-26-1-2233. Any opinions, findings, and conclusions or recommendations expressed in this material are those of the author(s) and do not necessarily reflect the views of the National Science Foundation or the Office of Naval Research.

\bibliography{colm2026_conference}
\bibliographystyle{colm2026_conference}

\newpage
\appendix
\section{Active Expert Comparison} \label{app:active_experts_comparison}

We evaluate performance when activating fewer than all 5 experts during router training and inference (Table~\ref{tab:active_experts}). Performance degrades substantially with fewer than 4 active experts, with the largest drops in math and code. However, activating 4 of 5 experts achieves nearly equivalent performance to the full model. This behavior is consistent with our use of coarse-grained, domain-level experts; splitting each expert into finer-grained sub-experts~\citep{he2024expertslicing} could improve routing flexibility and enable sparser activation without sacrificing performance. Table~\ref{tab:active_experts_comparison} compares \MODEL\ and BTX across active expert counts.

\begin{table}[t]
\begin{center}
\resizebox{\textwidth}{!}{
\begin{tabular}{clrrrrrrrr}
\toprule
\textbf{Active Experts} & \textbf{Method} & \textbf{Overall} & \textbf{Knowledge} & \textbf{Reasoning} & \textbf{Chat} & \textbf{Math} & \textbf{Code} & \textbf{Tool Use} & \textbf{Safety} \\
\midrule
\multirow{2}{*}{1} & \MODEL\ & 31.1 & 22.8 & 23.4 & 31.0 & 12.2 & 10.3 & 31.6 & 86.5 \\
                    & BTX    & 29.0 & 15.7 & 20.6 & 24.5 & 10.8 & 7.6 & 38.5 & 85.2 \\
\midrule
\multirow{2}{*}{2} & \MODEL\ & 38.8 & 27.5 & 28.1 & 37.1 & 23.4 & 19.3 & 44.5 & 91.9 \\
                    & BTX    & 38.7 & 24.1 & 26.6 & 35.3 & 26.0 & 20.8 & 43.6 & 94.6 \\
\midrule
\multirow{2}{*}{3} & \MODEL\ & 36.8 & 25.7 & 27.4 & 33.1 & 19.5 & 16.5 & 42.2 & 93.3 \\
                    & BTX    & 39.7 & 24.6 & 28.7 & 34.7 & 31.7 & 20.2 & 44.5 & 93.7 \\
\midrule
\multirow{2}{*}{4} & \MODEL\ & 48.2 & 28.5 & 31.0 & 37.5 & 55.1 & 48.5 & 42.6 & 94.5 \\
                    & BTX    & 45.1 & 24.2 & 29.7 & 35.0 & 59.0 & 28.2 & 45.5 & 93.9 \\
\midrule
\multirow{2}{*}{5} & \MODEL\ & 49.1 & 28.4 & 30.8 & 38.7 & 56.2 & 49.9 & 45.6 & 94.0 \\
                    & BTX    & 46.7 & 23.9 & 30.6 & 36.4 & 62.1 & 32.1 & 47.9 & 93.9 \\
\bottomrule
\end{tabular}
}
\end{center}
\caption{Comparison of \MODEL\ and BTX as the number of active experts varies. \MODEL\ outperforms BTX overall for all expert counts except n=3, with the gap largest at 4 active experts. Per-benchmark results are in Table~\ref{tab:detailed_active_comparison}.}
\label{tab:active_experts_comparison}
\end{table}

\section{Dataset Details} \label{app:dataset_details}

We list the full dataset composition for each domain expert below.

\tightparagraph{Math mid-training}
We use the math mid-training mix from FlexOlmo~\citep{shi2025flexolmoopenlanguagemodels}, which combines the Dolmino Math Mix~\citep{olmo2} and FineMath4+~\citep{finemath}.

\tightparagraph{Math SFT}
We use math-specific instruction data from Tulu~3~\citep{lambert2025tulu3pushingfrontiers}, which includes synthetically generated persona-driven math problems (Persona MATH, Persona GSM, Persona Algebra) as well as data from OpenMathInstruct~2~\citep{toshniwal2024openmathinstruct2acceleratingaimath} and NuminaMath-TIR~\citep{numinamath}.

\tightparagraph{Math RLVR}
We use the math verifiable reward data from OLMo~3~\citep{olmo2025olmo3}, which draws from Open-Reasoner-Zero~\citep{hu2025openreasonerzeroopensourceapproach}, DAPO-Math~\citep{dapomath}, AceReason-Math~\citep{acereason}, and OMEGA~\citep{omega}.

\tightparagraph{Code mid-training}
For code v2, we use the OLMo~3~\citep{olmo2025olmo3} code mid-training data, which includes StackEdu (quality-filtered code from The Stack with fill-in-the-middle formatting) and CraneCode (a permissively-licensed synthetic code dataset original to OLMo~3).

\tightparagraph{Code SFT}
For code v2, we use the OLMo~3~\citep{olmo2025olmo3} code SFT data (Dolci), which includes data from AceCoder~\citep{zeng2025acecoderacingcoderrl}, OpenCodeReasoning~\citep{ahmad2025opencodereasoningadvancingdatadistillation}, OpenThoughts~3~\citep{guha2025openthoughtsdatarecipesreasoning}, and Tulu~3 Persona Python~\citep{lambert2025tulu3pushingfrontiers}.

\tightparagraph{Code RLVR}
For code v2, we use the OLMo~3~\citep{olmo2025olmo3} code RLVR data, which includes prompts from AceCoder~\citep{zeng2025acecoderacingcoderrl} and other sources with execution-based verification.

\tightparagraph{Tool use SFT}
We use the Dolci tool use instruction data from OLMo~3~\citep{olmo2025olmo3}, which is largely original to that work, comprising synthetic function-calling trajectories (SimFC), science QA trajectories, and web search QA trajectories adapted from DR-Tulu~\citep{shao2025drtulureinforcementlearning}.

\tightparagraph{Safety SFT}
We use safety data following the OLMo~3~\citep{olmo2025olmo3} recipe, drawing from CoCoNot~\citep{coconot}, WildGuardMix~\citep{han2024wildguardopenonestopmoderation}, and WildJailbreak~\citep{jiang2024wildteamingscaleinthewildjailbreaks}.

\section{Training Details \& Hyperparameters} \label{app:training_details}

All experiments use OLMo 2 7B as the base model. Training was conducted on one to eight 8xH100 nodes with infiniband interconnect. We report hyperparameters for each training stage below.

\tightparagraph{Mid-training}
Mid-training uses a decaying learning rate schedule on large-scale domain corpora. Training is distributed across eight nodes (64 H100 GPUs).

\begin{tabular}{ll}
\toprule
Learning rate & 9e-4 \\
Warmup steps & 2,000 \\
Sequence length & 4096 \\
Max duration & 50B tokens \\
LR schedule & Cosine decay (from OLMo 2 anneal config) \\
Hardware & 8 nodes $\times$ 8 H100 GPUs \\
\bottomrule
\end{tabular}

\tightparagraph{Supervised Finetuning (SFT)}
Each domain expert is fine-tuned for 2 epochs on a mixture of domain-specific and general SFT data.

\begin{tabular}{ll}
\toprule
Learning rate & 1e-4 \\
Sequence length & 4,096 \\
Max duration & 2 epochs \\
Hardware & 2 nodes $\times$ 8 H100 GPUs \\
SFT data & Domain + general mix (see Table~\ref{tab:sft_mixing}) \\
\bottomrule
\end{tabular}

\tightparagraph{Reinforcement Learning (RLVR)}
We use GRPO with verifiable rewards for math and code.

\begin{tabular}{ll}
\toprule
Learning rate & 6e-7 \\
LR schedule & Constant \\
Max prompt length & 2,048 tokens \\
Response length & 4,096 tokens \\
Pack length & 7,168 tokens \\
Rollout samples per prompt & 8 \\
Unique prompts per rollout & 64 \\
Mini-batches & 4 \\
Hardware & 2 nodes $\times$ 8 H100 GPUs \\
Temperature & 1.0 \\
\bottomrule
\end{tabular}

\tightparagraph{Router Training}
After merging all experts, only the router parameters are trained.

\begin{tabular}{ll}
\toprule
Learning rate & 1e-4 \\
Sequence length & 2,048 \\
Max duration & 2 epochs \\
Training data & 5\% stratified SFT sample (all domains) \\
Hardware & 5 nodes $\times$ 8 H100 GPUs (1 per expert) \\
Frozen parameters & All non-router weights \\
\bottomrule
\end{tabular}

\section{Detailed Evaluation Results} \label{app:detailed_results}

We evaluate across seven domain categories. All evaluations use chat formatting with zero-shot chain-of-thought prompting unless otherwise noted.

\tightparagraph{Chat} We measure instruction-following and open-ended conversational quality using AlpacaEval~\citep{alpacaeval} (length-controlled win rate) and IFEval~\citep{zhou2023instructionfollowingevaluationlargelanguage} (prompt-level loose accuracy on verifiable instruction-following constraints).

\tightparagraph{Knowledge} We assess factual knowledge using MMLU~\citep{hendrycks2021measuringmassivemultitasklanguage} (zero-shot chain-of-thought), PopQA~\citep{mallen2023trustlanguagemodelsinvestigating} (15-shot, exact match), and SimpleQA~\citep{wei2024measuringshortformfactualitylarge} (F1).

\tightparagraph{Reasoning} We evaluate general reasoning with BBH~\citep{suzgun2022challengingbigbenchtaskschainofthought} (zero-shot chain-of-thought), GPQA~\citep{rein2023gpqagraduatelevelgoogleproofqa} (zero-shot chain-of-thought), ZebraLogic~\citep{lin2025zebralogicscalinglimitsllms} (constraint satisfaction and logical deduction), and AGIEval~\citep{zhong2023agievalhumancentricbenchmarkevaluating} (English subsets, zero-shot chain-of-thought).

\tightparagraph{Math} We use MATH~\citep{hendrycks2021measuringmathematicalproblemsolving} (zero-shot, exact match flex with \texttt{\textbackslash boxed\{\}} extraction) and GSM8K~\citep{gsm8k} (zero-shot, exact match flex).

\tightparagraph{Code} We measure code generation with HumanEval+~\citep{liu2023codegeneratedchatgptreally} (zero-shot chat, pass@1) and MBPP+~\citep{liu2023codegeneratedchatgptreally} (zero-shot chat, pass@1).

\tightparagraph{Tool Use} We evaluate function-calling and tool use with BFCL~\citep{patil2025bfcl} (accuracy across all subtasks).

\tightparagraph{Safety} We measure both proper refusal of harmful requests and avoidance of over-refusal using HarmBench~\citep{mazeika2024harmbenchstandardizedevaluationframework}, TrustLLM~\citep{huang2024trustllmtrustworthinesslargelanguage}, WildGuard~\citep{han2024wildguardopenonestopmoderation}, WildJailbreak~\citep{jiang2024wildteamingscaleinthewildjailbreaks} (benign subset to measure over-refusal), and Do Anything Now (DAN;~\citealp{shen2024donowcharacterizingevaluating}).

\vspace{1em}
Per-benchmark results for all models are reported in the tables below.

\newcommand{\rot}[1]{\rotatebox{90}{\small #1}}

\begin{table}[h]
\begin{center}
\resizebox{\textwidth}{!}{
\begin{tabular}{llrrrrrrrrrrrrr}
\toprule
& & & \multicolumn{6}{c}{\textit{Per-expert (2$\times$7B)}} & & \multicolumn{5}{c}{\textit{Baselines}} \\
\cmidrule(lr){4-9} \cmidrule(lr){11-15}
\textbf{Category} & \textbf{Benchmark}
  & \rot{Initial $M$ (7B)}
  & \rot{Math +SFT}
  & \rot{Math +RL}
  & \rot{Code +SFT}
  & \rot{Code +RL}
  & \rot{Tool Use}
  & \rot{Safety}
  & \rot{\MODEL\ 5$\times$7B}
  & \rot{Cont. post-train}
  & \rot{Merge (w/ mid)}
  & \rot{Merge (w/o mid)}
  & \rot{BTX 5$\times$7B}
  & \rot{Re-train (post)}
\\
\midrule
\multirow{2}{*}{Chat}
  & AlpacaEval & 20.0 & 11.2 & 12.1 & 10.0 & 8.9 & 10.1 & 9.3 & 11.4 & 8.6 & 0.0 & 13.8 & 7.8 & 17.0 \\
  & IFEval & 77.8 & 70.6 & 73.0 & 69.9 & 73.2 & 68.6 & 66.9 & 66.0 & 68.9 & 11.8 & 71.3 & 65.1 & 70.8 \\
\midrule
\multirow{3}{*}{Knowledge}
  & MMLU & 59.3 & 60.4 & 62.2 & 59.0 & 59.8 & 59.6 & 59.5 & 60.4 & 58.7 & 0.0 & 60.2 & 55.6 & 61.5 \\
  & PopQA & 21.7 & 21.6 & 20.9 & 22.3 & 21.5 & 21.2 & 21.9 & 20.5 & 18.4 & 0.0 & 22.7 & 13.6 & 20.2 \\
  & SimpleQA & 4.6 & 4.5 & 4.0 & 4.1 & 4.3 & 4.6 & 4.7 & 4.2 & 3.4 & 0.2 & 4.6 & 2.4 & 4.0 \\
\midrule
\multirow{4}{*}{Reasoning}
  & BBH & 39.9 & 37.7 & 38.8 & 37.6 & 38.6 & 36.2 & 36.3 & 39.1 & 36.7 & 7.5 & 38.8 & 38.3 & 40.3 \\
  & GPQA & 28.6 & 29.5 & 25.9 & 29.5 & 27.5 & 27.5 & 27.0 & 26.8 & 28.6 & 16.5 & 27.7 & 28.6 & 27.5 \\
  & ZebraLogic & 3.3 & 4.1 & 4.1 & 3.8 & 2.9 & 2.8 & 4.0 & 4.9 & 4.0 & 0.0 & 4.1 & 3.6 & 4.1 \\
  & AGIEval & 47.3 & 53.6 & 54.4 & 47.6 & 48.0 & 48.3 & 47.7 & 52.6 & 48.2 & 16.9 & 49.4 & 51.8 & 53.6 \\
\midrule
\multirow{2}{*}{Math}
  & MATH & 7.0 & 21.1 & 36.8 & 8.7 & 8.6 & 7.7 & 7.1 & 31.6 & 15.3 & 0.0 & 11.7 & 39.7 & 19.8 \\
  & GSM8K & 40.2 & 62.7 & 74.8 & 41.5 & 45.2 & 36.0 & 37.8 & 80.8 & 74.7 & 0.5 & 54.1 & 84.5 & 77.6 \\
\midrule
\multirow{2}{*}{Code}
  & HumanEval+ & 2.6 & 13.1 & 16.3 & 44.8 & 46.2 & 11.0 & 9.1 & 48.9 & 45.3 & 0.0 & 18.0 & 27.9 & 44.9 \\
  & MBPP+ & 21.0 & 27.8 & 27.8 & 50.7 & 54.6 & 22.9 & 22.3 & 50.8 & 40.4 & 0.8 & 32.2 & 36.3 & 42.3 \\
\midrule
Tool Use
  & BFCL & 25.3 & 21.6 & 19.8 & 19.7 & 19.8 & 46.4 & 21.1 & 45.6 & 40.9 & 19.7 & 19.7 & 47.9 & 45.3 \\
\midrule
\multirow{5}{*}{Safety}
  & HarmBench & 52.8 & 66.6 & 66.9 & 63.4 & 63.4 & 66.6 & 92.8 & 90.0 & 87.5 & 11.9 & 70.6 & 89.4 & 87.5 \\
  & TrustLLM & 21.8 & 55.0 & 59.0 & 69.2 & 62.5 & 67.5 & 94.5 & 96.0 & 92.5 & 5.8 & 65.8 & 94.5 & 86.0 \\
  & WildGuard & 68.4 & 77.3 & 77.6 & 78.6 & 75.8 & 78.6 & 99.1 & 98.9 & 99.2 & 19.0 & 84.8 & 98.1 & 99.6 \\
  & WildJailbreak & 100.0 & 98.4 & 100.0 & 99.6 & 99.2 & 99.2 & 88.4 & 88.0 & 89.2 & 2.4 & 98.8 & 88.8 & 98.8 \\
  & DAN & 13.3 & 66.0 & 73.3 & 75.7 & 75.7 & 83.7 & 98.3 & 97.0 & 97.0 & 6.7 & 72.7 & 98.7 & 92.3 \\
\bottomrule
\end{tabular}
}
\end{center}
\caption{Per-benchmark results for all models. Eval settings are described in \S4.3. Retraining (mid+post-train)$^\dagger$ is omitted for space; see Table~\ref{tab:stage_ablation}.}
\label{tab:detailed_results}
\end{table}

\begin{table}[h]
\begin{center}
\resizebox{\textwidth}{!}{
\begin{tabular}{llrrrrrrrrr}
\toprule
& & & \multicolumn{2}{c}{\textit{Math}} & \multicolumn{2}{c}{\textit{Code}} & \multicolumn{2}{c}{\textit{Tool Use}} & \multicolumn{2}{c}{\textit{Safety}} \\
\cmidrule(lr){4-5} \cmidrule(lr){6-7} \cmidrule(lr){8-9} \cmidrule(lr){10-11}
& & \rot{Initial $M$} & \rot{Domain only} & \rot{Mixed} & \rot{Domain only} & \rot{Mixed} & \rot{Domain only} & \rot{Mixed} & \rot{Domain only} & \rot{Mixed} \\
\midrule
\multirow{2}{*}{Chat}
  & AlpacaEval & 20.0 & 1.2 & 11.2 & 4.0 & 10.0 & 3.7 & 10.1 & 8.7 & 9.3 \\
  & IFEval & 77.8 & 30.1 & 70.6 & 48.8 & 69.9 & 40.5 & 68.6 & 32.2 & 66.9 \\
\midrule
\multirow{3}{*}{Knowledge}
  & MMLU & 59.3 & 48.1 & 60.4 & 53.3 & 59.0 & 42.8 & 59.6 & 56.1 & 59.5 \\
  & PopQA & 21.7 & 15.7 & 21.6 & 21.2 & 22.3 & 10.1 & 21.2 & 14.8 & 21.9 \\
  & SimpleQA & 4.6 & 2.5 & 4.5 & 3.5 & 4.1 & 3.0 & 4.6 & 4.0 & 4.7 \\
\midrule
\multirow{4}{*}{Reasoning}
  & BBH & 39.9 & 25.4 & 37.7 & 24.5 & 37.6 & 16.0 & 36.2 & 26.6 & 36.3 \\
  & GPQA & 28.6 & 26.1 & 29.5 & 29.9 & 29.5 & 22.8 & 27.5 & 24.1 & 27.0 \\
  & ZebraLogic & 3.3 & 0.8 & 4.1 & 3.3 & 3.8 & 0.1 & 2.8 & 1.5 & 4.0 \\
  & AGIEval & 47.3 & 42.8 & 53.6 & 43.2 & 47.6 & 41.9 & 48.3 & 45.8 & 47.7 \\
\midrule
\multirow{2}{*}{Math}
  & MATH & 7.0 & 21.2 & 21.1 & 6.1 & 8.7 & 4.1 & 7.7 & 4.7 & 7.1 \\
  & GSM8K & 40.2 & 46.0 & 62.7 & 26.3 & 41.5 & 21.1 & 36.0 & 36.1 & 37.8 \\
\midrule
\multirow{2}{*}{Code}
  & HumanEval+ & 2.6 & 20.4 & 13.1 & 49.9 & 44.8 & 6.5 & 11.0 & 10.8 & 9.1 \\
  & MBPP+ & 21.0 & 25.2 & 27.8 & 50.0 & 50.7 & 6.8 & 22.9 & 21.5 & 22.3 \\
\midrule
Tool Use
  & BFCL & 25.3 & 19.7 & 21.6 & 20.5 & 19.7 & 44.9 & 46.4 & 19.7 & 21.1 \\
\midrule
\multirow{5}{*}{Safety}
  & HarmBench & 52.8 & 32.2 & 66.6 & 61.9 & 63.4 & 89.1 & 66.6 & 99.7 & 92.8 \\
  & TrustLLM & 21.8 & 13.5 & 55.0 & 52.0 & 69.2 & 58.8 & 67.5 & 99.0 & 94.5 \\
  & WildGuard & 68.4 & 53.3 & 77.3 & 72.6 & 78.6 & 75.0 & 78.6 & 98.8 & 99.1 \\
  & WildJailbreak & 100.0 & 95.6 & 98.4 & 98.0 & 99.6 & 62.8 & 99.2 & 87.6 & 88.4 \\
  & DAN & 13.3 & 21.7 & 66.0 & 42.7 & 75.7 & 79.0 & 83.7 & 100.0 & 98.3 \\
\bottomrule
\end{tabular}
}
\end{center}
\caption{Per-benchmark results for SFT data mixing ablation (Table~\ref{tab:sft_mixing}).}
\label{tab:detailed_sft_mixing}
\end{table}

\begin{table}[h]
\begin{center}
\small
\begin{tabular}{llrrrrr}
\toprule
& & \rot{Initial $M$} & \rot{+ Math (2$\times$7B)} & \rot{+ Code (3$\times$7B)} & \rot{+ Tool Use (4$\times$7B)} & \rot{+ Safety (5$\times$7B)} \\
\midrule
\multirow{2}{*}{Chat}
  & AlpacaEval & 20.0 & 12.0 & 8.7 & 10.4 & 11.4 \\
  & IFEval & 77.8 & 74.9 & 68.4 & 70.8 & 66.0 \\
\midrule
\multirow{3}{*}{Knowledge}
  & MMLU & 59.3 & 59.8 & 59.1 & 60.3 & 60.4 \\
  & PopQA & 21.7 & 21.5 & 21.8 & 21.3 & 20.5 \\
  & SimpleQA & 4.6 & 4.1 & 4.2 & 4.6 & 4.2 \\
\midrule
\multirow{4}{*}{Reasoning}
  & BBH & 39.9 & 39.4 & 38.6 & 38.7 & 39.1 \\
  & GPQA & 28.6 & 31.7 & 25.7 & 29.5 & 26.8 \\
  & ZebraLogic & 3.3 & 4.0 & 4.5 & 3.9 & 4.9 \\
  & AGIEval & 47.3 & 52.8 & 53.0 & 50.8 & 52.6 \\
\midrule
\multirow{2}{*}{Math}
  & MATH & 7.0 & 40.0 & 37.3 & 32.4 & 31.6 \\
  & GSM8K & 40.2 & 75.8 & 79.8 & 80.6 & 80.8 \\
\midrule
\multirow{2}{*}{Code}
  & HumanEval+ & 2.6 & 13.1 & 51.2 & 49.4 & 48.9 \\
  & MBPP+ & 21.0 & 27.2 & 45.0 & 47.2 & 50.8 \\
\midrule
Tool Use
  & BFCL & 25.3 & 21.0 & 19.9 & 45.8 & 45.6 \\
\midrule
\multirow{5}{*}{Safety}
  & HarmBench & 52.8 & 62.5 & 59.4 & 62.8 & 90.0 \\
  & TrustLLM & 21.8 & 50.8 & 60.2 & 63.3 & 96.0 \\
  & WildGuard & 68.4 & 77.2 & 78.1 & 77.6 & 98.9 \\
  & WildJailbreak & 100.0 & 98.8 & 98.8 & 98.8 & 88.0 \\
  & DAN & 13.3 & 58.3 & 65.7 & 72.0 & 97.0 \\
\bottomrule
\end{tabular}
\end{center}
\caption{Per-benchmark results for expert interference (Table~\ref{tab:expert_interference}).}
\label{tab:detailed_interference}
\end{table}

\begin{table}[h]
\begin{center}
\resizebox{\textwidth}{!}{
\begin{tabular}{llrrrrrrrr}
\toprule
& & \multicolumn{4}{c}{\textit{Math}} & \multicolumn{4}{c}{\textit{Code}} \\
\cmidrule(lr){3-6} \cmidrule(lr){7-10}
& & \rot{Math v1 (SFT only)} & \rot{Math v2 (SFT + RL)} & \rot{\MODEL\ w/ Math v1} & \rot{\MODEL\ w/ Math v2} & \rot{Code v1} & \rot{Code v2} & \rot{\MODEL\ w/ Code v1} & \rot{\MODEL\ w/ Code v2} \\
\midrule
\multirow{2}{*}{Chat}
  & AlpacaEval & 11.2 & 12.0 & 10.6 & 9.6 & 9.6 & 9.9 & 10.0 & 9.6 \\
  & IFEval & 70.6 & 74.9 & 65.4 & 66.7 & 68.0 & 72.3 & 63.6 & 66.7 \\
\midrule
\multirow{3}{*}{Knowledge}
  & MMLU & 60.4 & 59.8 & 60.2 & 60.0 & 59.5 & 59.6 & 60.0 & 60.0 \\
  & PopQA & 21.6 & 21.5 & 22.0 & 21.8 & 22.4 & 21.2 & 21.9 & 21.8 \\
  & SimpleQA & 4.5 & 4.1 & 4.5 & 4.0 & 4.6 & 4.5 & 4.5 & 4.0 \\
\midrule
\multirow{4}{*}{Reasoning}
  & BBH & 37.7 & 39.4 & 38.2 & 38.7 & 37.5 & 37.1 & 39.7 & 38.7 \\
  & GPQA & 29.5 & 31.7 & 30.8 & 28.3 & 28.3 & 27.5 & 28.8 & 28.3 \\
  & ZebraLogic & 4.1 & 4.0 & 3.6 & 4.5 & 3.1 & 3.6 & 4.3 & 4.5 \\
  & AGIEval & 53.6 & 52.8 & 52.3 & 51.1 & 48.0 & 47.5 & 52.5 & 51.1 \\
\midrule
\multirow{2}{*}{Math}
  & MATH & 21.1 & 40.0 & 21.3 & 33.4 & 8.0 & 7.8 & 32.9 & 33.4 \\
  & GSM8K & 62.7 & 75.8 & 65.2 & 79.6 & 39.8 & 43.8 & 80.2 & 79.6 \\
\midrule
\multirow{2}{*}{Code}
  & HumanEval+ & 13.1 & 13.1 & 51.1 & 50.5 & 25.1 & 49.2 & 29.0 & 50.5 \\
  & MBPP+ & 27.8 & 27.2 & 48.1 & 46.0 & 34.2 & 53.4 & 37.9 & 46.0 \\
\midrule
Tool Use
  & BFCL & 21.6 & 21.0 & 45.0 & 45.5 & 21.5 & 19.7 & 45.9 & 45.5 \\
\midrule
\multirow{5}{*}{Safety}
  & HarmBench & 66.6 & 62.5 & 88.8 & 89.1 & 67.5 & 62.8 & 90.9 & 89.1 \\
  & TrustLLM & 55.0 & 50.8 & 93.5 & 93.8 & 65.0 & 73.0 & 94.2 & 93.8 \\
  & WildGuard & 77.3 & 77.2 & 98.8 & 98.8 & 78.1 & 78.2 & 98.5 & 98.8 \\
  & WildJailbreak & 98.4 & 98.8 & 87.6 & 90.8 & 98.4 & 98.8 & 90.0 & 90.8 \\
  & DAN & 66.0 & 58.3 & 96.0 & 96.7 & 74.0 & 70.0 & 97.7 & 96.7 \\
\bottomrule
\end{tabular}
}
\end{center}
\caption{Per-benchmark results for modular upgrades (Table~\ref{tab:modular_upgrade}).}
\label{tab:detailed_upgrades}
\end{table}

\begin{table}[h]
\begin{center}
\resizebox{\textwidth}{!}{
\begin{tabular}{llrrrrrrrrrr}
\toprule
& & \multicolumn{5}{c}{\textit{\MODEL}} & \multicolumn{5}{c}{\textit{BTX}} \\
\cmidrule(lr){3-7} \cmidrule(lr){8-12}
& & \rot{1} & \rot{2} & \rot{3} & \rot{4} & \rot{5} & \rot{1} & \rot{2} & \rot{3} & \rot{4} & \rot{5} \\
\midrule
\multirow{2}{*}{Chat}
  & AlpacaEval & 3.7 & 8.9 & 8.4 & 11.3 & 9.6 & 2.0 & 7.2 & 8.3 & 7.9 & 7.8 \\
  & IFEval & 58.2 & 65.2 & 57.9 & 63.6 & 66.7 & 47.0 & 63.4 & 61.2 & 62.1 & 65.1 \\
\midrule
\multirow{3}{*}{Knowledge}
  & MMLU & 48.5 & 57.8 & 52.7 & 59.2 & 60.0 & 36.0 & 53.6 & 54.3 & 55.2 & 55.6 \\
  & PopQA & 17.0 & 20.9 & 20.8 & 22.2 & 21.8 & 9.7 & 15.4 & 16.4 & 14.5 & 13.6 \\
  & SimpleQA & 2.9 & 3.7 & 3.6 & 4.2 & 4.0 & 1.4 & 3.3 & 3.0 & 3.0 & 2.4 \\
\midrule
\multirow{4}{*}{Reasoning}
  & BBH & 31.4 & 37.0 & 34.6 & 38.1 & 38.7 & 26.8 & 35.8 & 37.2 & 37.5 & 38.3 \\
  & GPQA & 25.9 & 26.6 & 29.2 & 28.1 & 28.3 & 22.1 & 24.8 & 28.3 & 26.1 & 28.6 \\
  & ZebraLogic & 0.3 & 3.2 & 2.3 & 5.0 & 4.5 & 1.8 & 2.5 & 4.5 & 5.0 & 3.6 \\
  & AGIEval & 35.9 & 45.9 & 43.6 & 52.9 & 51.1 & 31.6 & 43.3 & 44.6 & 50.2 & 51.8 \\
\midrule
\multirow{2}{*}{Math}
  & MATH & 4.2 & 6.9 & 5.7 & 32.2 & 33.4 & 3.9 & 8.4 & 11.1 & 35.8 & 39.7 \\
  & GSM8K & 20.2 & 39.8 & 33.4 & 77.9 & 79.6 & 17.7 & 43.5 & 52.3 & 82.2 & 84.5 \\
\midrule
\multirow{2}{*}{Code}
  & HumanEval+ & 6.3 & 12.7 & 12.0 & 47.9 & 50.5 & 4.5 & 15.4 & 14.9 & 22.9 & 27.9 \\
  & MBPP+ & 14.4 & 25.9 & 21.1 & 49.1 & 46.0 & 10.7 & 26.3 & 25.4 & 33.5 & 36.3 \\
\midrule
Tool Use
  & BFCL & 31.6 & 44.5 & 42.2 & 42.6 & 45.5 & 38.5 & 43.6 & 44.5 & 45.5 & 47.9 \\
\midrule
\multirow{5}{*}{Safety}
  & HarmBench & 80.6 & 90.6 & 87.5 & 91.9 & 89.1 & 83.8 & 91.2 & 91.6 & 88.1 & 89.4 \\
  & TrustLLM & 84.5 & 93.0 & 93.5 & 95.0 & 93.8 & 82.5 & 94.2 & 94.2 & 94.8 & 94.5 \\
  & WildGuard & 92.9 & 99.1 & 98.5 & 99.3 & 98.8 & 96.1 & 98.8 & 97.9 & 98.5 & 98.1 \\
  & WildJailbreak & 84.0 & 84.0 & 89.2 & 88.0 & 90.8 & 72.8 & 88.8 & 86.4 & 90.0 & 88.8 \\
  & DAN & 90.3 & 92.7 & 98.0 & 98.3 & 96.7 & 91.0 & 99.7 & 98.3 & 98.3 & 98.7 \\
\bottomrule
\end{tabular}
}
\end{center}
\caption{Per-benchmark comparison of \MODEL\ vs BTX across active expert counts (Table~\ref{tab:active_experts_comparison}).}
\label{tab:detailed_active_comparison}
\end{table}

\end{document}